\documentclass[10pt, conference, compsocconf]{IEEEtran}
\ifCLASSOPTIONcompsoc
  \usepackage[nocompress]{cite}
\else
  \usepackage{cite}
\fi

\ifCLASSINFOpdf
\else
\fi

\usepackage{mathtools}
\usepackage{amsmath}
\usepackage{amssymb}
\usepackage{bm}
\usepackage{subcaption}
\usepackage{esvect}

\usepackage{tikz}
\usetikzlibrary{arrows}
\usetikzlibrary{calc}
\usetikzlibrary{shapes.geometric}

\usetikzlibrary{positioning}
\usepackage{tkz-graph}

\usepackage{times}
\usepackage{url}
\usepackage{latexsym}
\usepackage{amsmath}
\usepackage[ruled,linesnumbered]{algorithm2e}
\usepackage{graphicx}

\usepackage{multirow}
\usepackage[font={small}]{caption}

\usepackage{float}
\floatstyle{plaintop}
\restylefloat{table}

\SetAlFnt{\small}

\begin{document}
%
% paper title
% Titles are generally capitalized except for words such as a, an, and, as,
% at, but, by, for, in, nor, of, on, or, the, to and up, which are usually
% not capitalized unless they are the first or last word of the title.
% Linebreaks \\ can be used within to get better formatting as desired.
% Do not put math or special symbols in the title.
\title{Product Function Need Recognition via Semi-supervised Attention Network}
\author{
    \IEEEauthorblockN{Hu Xu\IEEEauthorrefmark{1}, Sihong Xie\IEEEauthorrefmark{2}, Lei Shu\IEEEauthorrefmark{1}, Philip S. Yu\IEEEauthorrefmark{1}\IEEEauthorrefmark{3}}
    \IEEEauthorblockA{\IEEEauthorrefmark{1}Department of Computer Science, University of Illinois at Chicago, Chicago, IL, USA}
    \IEEEauthorblockA{\IEEEauthorrefmark{2}Department of Computer Science and Engineering, Lehigh University, Bethlehem, PA, USA}
    \IEEEauthorblockA{\IEEEauthorrefmark{3}Institute for Data Science, Tsinghua University, Beijing, China}
    \IEEEauthorblockA{hxu48@uic.edu, sxie@cse.lehigh.edu, lshu3@uic.edu, psyu@uic.edu}
}

\maketitle

\begin{abstract}
Functionality is of utmost importance to customers when they purchase products.
However, it is unclear to customers whether a product can really satisfy their needs on functions.
Further, missing functions may be intentionally hidden by the manufacturers or the sellers.
As a result, a customer needs to spend a fair amount of time before purchasing or just purchase the product on his/her own risk.

In this paper, we first identify a novel QA corpus that is dense on product functionality information \footnote{The annotated corpus can be found at \url{https://www.cs.uic.edu/~hxu/}.}.
We then design a neural network called Semi-supervised Attention Network (SAN) to discover product functions from questions.
This model leverages unlabeled data as contextual information to perform semi-supervised sequence labeling.
We conduct experiments to show that the extracted function have both high coverage and accuracy, compared with a wide spectrum of baselines.    
\end{abstract}

\begin{IEEEkeywords}
product function need recognition; semi-supervised learning; deep learning; attention

\end{IEEEkeywords}

\section{Introduction}
\label{intro}
Functionality is a fundamental concern for customers when they decide to buy a new product. 
From \textit{customers}' perspective, before they purchase a product, it is natural for them to ask what the to-be-purchased one can do and cannot do.
From \textit{sellers}' perspective, selling fully-functioned products can increase sales,
and yet selling products with missing functions can lead to catastrophic customer dissatisfaction.
From \textit{manufacturers}' perspective, missing functions reported by customers can help improve their products.
In marketing, the term \textit{product} is defined as ``anything that can be offered to a market for attention, acquisition, use or consumption that might satisfy a want or need'' \cite{kotler2010principles}. 
It is crucial to ensure that the functions of a product can satisfy customers' needs.
Therefore, conveying the information about functions successfully to customers is important for both manufacturers and sellers.

\begin{table}[th]
\footnotesize
\centering
\scalebox{1.05}{
\begin{tabular}{ l | l }
\hline
\multicolumn{2}{ l }{ 
\begin{tabular}[t]{@{}l@{}} 
Apple 13 " MacBook Pro \\
(2.5GHz Intel Core i5, 4GB RAM, 500GB HDD) 
\end{tabular}
}
\\
\hline
Q: & Can I \underline{\textbf{use}} this \underline{\textbf{for} video editing}\\
A: & No, it does not support Google Play. \\
\hline
Q: & Can I \underline{\textbf{make} video calls} to other non Apple computers ? ? \\
A: & yes you can if they have Skype , Tango , or oovoo\\
\hline
Q: & Will it be \underline{\textbf{useful for} music production} ? \\
A: & 
\begin{tabular}[t]{@{}l@{}} 
I have not used it for music production ; \\
however , I believe that it would be and have \\
several friends who use it specifically for that purpose .\\
\end{tabular} \\
\hline
Q: & Can I \underline{\textbf{use} Microsoft Office} on this MacBook Pro ? \\
A: & 
\begin{tabular}[t]{@{}l@{}} 
You can but maybe you wo n't want to . \\
The current Apple MacBook Pro is shipping with the \\
Mavericks operating system , which includes Pages , \\
Numbers , and Keynote at no cost .
\end{tabular} \\
\hline
\end{tabular}
}
\caption{A few QA pairs for a laptop: function expressions are underlined with function words (e.g., verbs, adjectives or prepositions) bolded.}
\label{table:sample}
\end{table}

In e-commerce platforms, one issue to convey such information is that products cannot be physically presented to customers before purchasing.
To overcome such limitation, many alternative approaches are deployed, i.e., using descriptions, pictures, and videos.
However, detailed functionality information may not be readily available for the following reasons. 
1) The cost of testing functions multiplied by a large number of products can be extremely high.
For example, it is impossible to test so many PCs whether they can run specific high-performance PC games.
2) Some missing functions are deliberatively hidden from descriptions by sellers to avoid hurting sales. 

Fortunately,
functionality information can be exchanged between customers and sellers via online platforms, such as forums and community QA.
This allows us to adopt an NLP-based approach to automatically sense and harvest product functions on a large scale.
We formulate a novel text mining task called Function Need Recognition (or FNR for short). 
A function need is defined as a sequence of words indicate a function expression (e.g., ``make video calls'').
In this paper, we only focus on product function needs and leave satisfiability issues (e.g., whether a product can ``make video calls'') to future work 
\footnote{A comprehensive study of product function satisfiability can be found at AAAI-2018 \cite{Xu2018pro}. }.

This task is non-trivial and the following challenges have to be addressed.
First, to ensure extraction quality, corpora that are dense and accurate in product functionality information are preferred.
To the best of our knowledge, there is no existing study on such a corpus to meet these requirements.
Second, the number of function needs can be unlimited.
How to ensure unexpected function needs can be detected is important. 

We address the challenges by first identify and annotate a high-quality corpus.
In particular, Amazon.com
allows potential consumers to communicate with existing product owners or sellers regarding product functions via Product Community Question Answering (PCQA for short).
Four (4) QA pairs talking about a laptop sold on Amazon are shown in Table \ref{table:sample}. 
Observe that the name of target product (to-be-purchased) can be identified using the metadata of the target product.
But 4 function needs (``use for video editing'', ``make video calls'', ``useful for music production'', and ``use Microsoft Office'') should be identified from the questions. 

Given the corpus,
we then formulate the problem as a sequence labeling task on questions.
We propose a deep sequence labeling model called Semi-supervised Attention Network (SAN) to solve this problem.
The key property of SAN is to use attention mechanism to summarize unlabeled data as side information for short labeled questions.
For example, let us assume only the 1st question is in the labeled data and all other 3 questions are in unlabeled data.
Then words like ``use'' or ``video'' in other 3 questions can serve as side information to help identify that ``use for video editing'' is a function.
Also, another advantage of using unlabeled data is that the embeddings of words do not appear in labeled data can still be tuned during training. 
To the best of our knowledge, this is the first attempt to use attention mechanism in a semi-supervised setting. 

\section{Model and Preliminary}
\subsection{Model Overview}
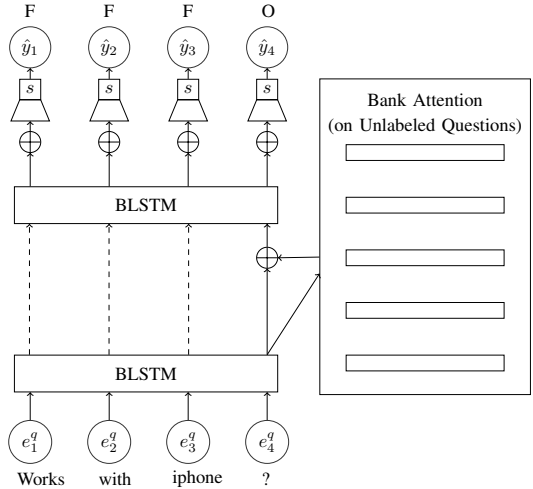
\begin{figure}[tp]
    \scalebox{0.7}{
    \centering
     \begin{tikzpicture}
    [cross/.style={path picture={ 
            \draw[black](path picture bounding box.south) -- (path picture bounding box.north) (path picture bounding box.west) -- (path picture bounding box.east);
        }}]
        
        \tikzstyle{crl}=[circle, minimum size = 7mm, draw =black!80, fill=white, node distance = 20mm]
        \tikzstyle{ucrl}=[circle, minimum size = 3mm, draw =black!80, fill=white, node distance = 20mm]
        \tikzstyle{box}=[minimum size = 7mm, draw =black!80, node distance = 20mm]
        \tikzstyle{concat} = [draw,circle,cross,minimum width=4mm]
        \tikzstyle{dense} = [trapezium, trapezium angle=67.5, draw,minimum height=4mm]
        \tikzstyle{softmax} = [minimum size = 1mm, draw =black!]
        \tikzstyle{sblstm}=[rectangle, draw, fill=white,minimum width=50mm, minimum height=7mm]
        \tikzstyle{ulstm}=[rectangle, draw, fill=white,minimum width=30mm, minimum height=3mm]
        \tikzstyle{atn}=[rectangle, draw,  fill=white,minimum width=40mm, minimum height=60mm]
        \tikzstyle{box}=[minimum size = 7mm, draw =black!80, node distance = 20mm]
        
        \node[text width=5mm] (q_i1) at (1.5*1-18,-2.2) {Works};
        \node[text width=4mm] (q_i2) at (1.5*2-18,-2.2) {with};
        \node[text width=6mm] (q_i3) at (1.5*3-18,-2.2) {iphone};
        \node[text width=2mm] (q_i4) at (1.5*4-18,-2.2) {?};
        
        \foreach \name in {1,...,4}{
            \node[crl] (q\name) at (1.5*\name-18,-1.5) {$e^q_\name$};            
            \node[concat] (q_2c\name) at (1.5*\name-18,4.2){};            
            \node[dense] (q_d\name) at (1.5*\name-18,4.8){};            
            \node[softmax] (q_s\name) at (1.5*\name-18,5.2){$s$};            
            \node[crl] (y\name) at (1.5*\name-18, 6.0) {$\hat{y}_\name$};
        }
        \node[text width=2mm] (q_o1) at (1.5*1-18,6.7) {F};
        \node[text width=2mm] (q_o2) at (1.5*2-18,6.7) {F};
        \node[text width=2mm] (q_o3) at (1.5*3-18,6.7) {F};
        \node[text width=2mm] (q_o4) at (1.5*4-18,6.7) {O};
        \node[sblstm] (lstm_q1) at (-8.3-6,-0.2)  {BLSTM} ;
        \node[sblstm] (lstm_q2) at (-8.3-6,3)  {BLSTM} ;
        \node[concat] (c) at (-12, 2) {};
        \node[atn] (atn) at (-9, 2.4) {};
        \node[] (atn_text) at (-9, 5) {Bank Attention};
        \node[] (atn_text) at (-9, 4.5) {(on Unlabeled Questions)};
        
        \foreach \h in {1,...,5}{
                \node[ulstm] (lstm_a\h) at (-9,-1+1*\h)  {} ;
        }        
        \foreach \h in {1,...,4}
        {
            \path (q_2c\h) edge [->] (q_d\h);
            \path (q_s\h) edge [->] (y\h);        
            \draw[->] (lstm_q2.north -| q_2c\h.south) -- (q_2c\h.south) ;            
            \draw[<-] (lstm_q1.south -| q\h.north) -- (q\h.north);
        }
        \draw[->] ([yshift=-11]atn.west) -- (c.east);
        \draw[->] (lstm_q1.north -| c.south) -- (c.south) ;
        \draw[<-] (lstm_q2.south -| c.north) -- (c.north) ;
        \draw[->] (lstm_q1.north -| c.south) -- ([yshift = -20]atn.west) ;
        
        \foreach \h in {1,...,3}
        {
            \draw[transform canvas={xshift=43*\h-106},dashed,->] ([xshift = 0]lstm_q1) -- (lstm_q2) ;
        }                
        \end{tikzpicture}
        }
        \caption{Semi-supervised Attention Network (SAN): the bottom 4 words are an input (labeled) question. They are labeled as $F$, $F$, $F$, $O$, indicating ``Works with iphone'' is a function expression. On the right is bank attention on unlabeled questions (sample questions are omitted). }
        \label{fig:fr}
    \end{figure}

We briefly introduce the proposed Semi-supervised Attention Network (SAN) in this section.
The idea of the network is to couple RNN-based sequence labeling network with attention on unlabeled data.
The proposed network is illustrated in Fig. \ref{fig:fr}. 
The left side can be viewed as a supervised sequence labeling model. 
It reads in a (labeled) question $\mathbf{x}^q$ and outputs label sequence $\mathbf{y}=(y_1, \dots, y_t, \dots, y_{T_q})$, where $y_t=l \in L=\{F, O\}$. 
The right side is the semi-supervised part.
A few unlabeled questions $U=\{\mathbf{x}^{u_1}, \dots, \mathbf{x}^{u_n}, \dots, \mathbf{x}^{u_{|U|}}\}$ are fed into a bank of BLSTMs (Bidirectional Long Short-Term Memory \cite{hochreiter1997long,schuster1997bidirectional}, one for each unlabeled question) with attentions (called bank attention).
The attended results are served as side information for the (labeled) question.
The key point here is, given a labeled question, we need to learn the weights on how to attend (or read) unlabeled questions.
Note that both supervised and semi-supervised parts share the same embedding layer.
This also gives the opportunity to tune embeddings of words not appear in the labeled questions.
Such a tuning is impossible in supervised settings.
All unlabeled questions share the same weights for their BLSTM layers (not shown in the figure). 
%Later, the outputs of unlabeled BLSTMs are attended by each position of the labeled BLSTM. 
After each word in the labeled question obtains the side formation, 
we feed the augmented labeled question into another BLSTM layer.
Then we generate label sequence $\mathbf{y}$ via a softmax layer. 
Overall, the labeled question can leverage unlabeled questions to decide the output labels in an end-to-end manner. 

\subsection{Preliminary}
\textbf{Embedding Layer}
We pair each labeled question $\mathbf{x}^q$ with a few unlabeled questions $U=\mathbf{x}^{u_{1:|U|}}$ (for both the training data and the test data). 
Unlabeled questions are similar questions from the same category as the labeled question returned by a search engine.
Let the sequence $\mathbf{x}^q=(x_1^q, \dots, x_{T_q}^q )$ and $\mathbf{x}^{u_n}=(x_1^{u_n}, \dots, x_{T_{u_n}}^{u_n} )$ denote the labeled question and the $n$-th unlabeled question, respectively.
Here $T_q$ and $T_{u_n}$ denote their respective lengths. 
When a question contains multiple sentences, we concatenate them into a single sequence.
We separate the sentences by a special token \textit{EOS}. 
We set $T_q=T_{u_{1:n}}=40$, which covers 99.5\% of lengths of labeled questions. 
Questions longer (shorter) than $40$ words are truncated (padded with zeros).
%Each input word in a question is a one-hot vector: e.g., $x_t^q \in \{0, 1\}^V$, where $V$ is the vocabulary size. 
We can view $\mathbf{x}^q$ ($\mathbf{x}^{u_n}$, resp.) as a matrix of one-hot column vectors. 
$\mathbf{x}^q$ is later transformed into embedded representation $\boldsymbol{e}^q$ ($\boldsymbol{e}^{u_n}$, resp.).
%This transform is done via a multiplication with a word embedding matrix $W_e$: $\boldsymbol{e}^q=W_e \mathbf{x}^q$.
%Here $W_e \in \mathbb{R}^{d_e \times V}$, and $d_e$ is the dimension (we set it to 100) of word vectors.
We pre-train the word embedding via skip-gram model \cite{mikolov2013distributed}. 
Then we fine-tune the embeddings when optimizing the proposed model. 
%Note that the unlabeled questions share the same trainable embedding matrix as the labeled questions.
%So the semi-supervised part also helps to learn word embeddings for words may not appear in the labeled questions.

\textbf{BLSTM Layer}
The embedded question sequences ($\boldsymbol{e}^q$ and $\boldsymbol{e}^{u_{1:|U|}}$) are fed into the labeled BLSTM and the unlabeled BLSTMs, respectively. 
We use $\boldsymbol{h}^{q, 1}$ and $\boldsymbol{h}^{u_{1:|U|}}$ to denote the outputs of these BLSTM layers for the labeled question and unlabeled questions, respectively.
We show important notations in Table \ref{table:not}, which is used in the next section.

\section{Semi-supervised Attention Network}

\begin{table}
%\footnotesize
\centering
\scalebox{0.8}{
\begin{tabular}{ l | l }
\hline
Notation & Explanation \\
\hline
$\boldsymbol{h}^{q, d}$ & 
\begin{tabular}[t]{@{}l@{}} 
The $d$-th hidden representation of the labeled 
\\question $q$ ($d=1, 2, 3$)
\end{tabular}\\
\hline
$\boldsymbol{h}^{u_n}$ & 
\begin{tabular}[t]{@{}l@{}} 
Hidden representations of the $n$-th unlabeled\\
question $u_n$
\end{tabular}\\
\hline
$t$ & 
The $t$-th word in the labeled question \\
\hline
$v$ & 
The $v$-th word in an unlabeled question \\
\hline
$r$ & 
\begin{tabular}[t]{@{}l@{}} 
Indicator of transformed representation \\
for the labeled question $q$ \\
\end{tabular}\\
\hline
$k, k'$ & 
\begin{tabular}[t]{@{}l@{}} 
Indicators of transformed represention \\
for the unlabeled question $u_n$ \\
\end{tabular}\\
\hline
$\alpha_{t, v}^{q, u_n}$ & 
\begin{tabular}[t]{@{}l@{}} 
Level 1 attention weights for the \\
$t$-th word in $q$ on the $v$-th word in $u_n$.
\end{tabular}\\
\hline
$\alpha_t^{q, u_n}$ & 
\begin{tabular}[t]{@{}l@{}} 
Level 2 attention weights for the \\
$t$-th word in $q$ on $u_n$
\end{tabular}\\
\hline
$h_t^{q, u_n}$ & 
\begin{tabular}[t]{@{}l@{}} 
Level 1 attended representation: \\
the $t$-th word in $q$ attends on unlabeled question $u_n$
\end{tabular}\\
\hline
$s_t^q$ & 
\begin{tabular}[t]{@{}l@{}} 
Level 2 attended representation: the $t$-th word\\ 
in $q$ attends on all $U=u_{1:n}$
\end{tabular}\\
\hline
\end{tabular}
}
\caption{Notations}
\label{table:not}
\end{table}

\subsection{Bank Attention}
\begin{figure}
\centering
\scalebox{0.8}{
\includegraphics[width=3in]{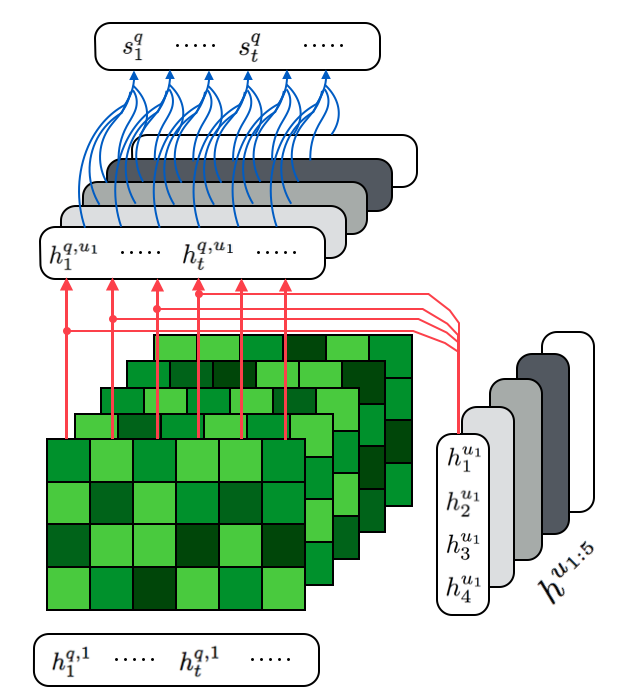}
}
\caption{Bank Attention: the $t$-th word representation $h_t^{q, 1}$ obtains its side information $s_t^q$ from multiple unlabeled questions such as the 1st unlabeled question $h_{1:4}^{u_1}$. The red arrows indicate level 1 attention among different words in one unlabeled question (we omit the arrows for the other 4 questions). The blue arrows indicate level 2 attention among multiple representations of unlabeled questions.}
\label{fig:bkattn}
\end{figure}

The key point of SAN is to leverage attention mechanism for semi-supervised learning. 
%In traditional semi-supervised learning \cite{zhu2005semi,salimans2016improved,nguyen2016semi} the labeled data and unlabeled data are passed via the same input but optimized for different objectives.
%Instead, we leverage unlabeled data to provide side information for labeled data. 
%Attention mechanism \cite{NIPS2010_4089,denil2012learning} is popular due to its capability of modeling variable length memories rather than the fixed-length memory. 
We utilize attention mechanism to synthesize side information from unlabeled data for each word in a labeled question.
The idea is that words in unlabeled data may have useful information for sequence labeling when they talk about similar products.
%So the performance to label a short question can be improved. 
We introduce a hierarchical attention mechanism.
As traditional attention mechanism, we let each word in a labeled question to attend a word in an unlabeled question.
This is level 1 attention.
On the higher level, 
we pair a labeled question with multiple related unlabeled questions. 
Note that different questions may not equally contribute side information to the labeled question. 
So we allow one word in the labeled question to attend on the results of level 1 attention on multiple questions.
We use the term \emph{bank attention} to refer to one word in a labeled question hierarchically attending to unlabeled questions.
The details are shown in Fig. \ref{fig:bkattn}.

We try to get the side information for the $t$-th word in the labeled question.
We first transform the word representations of the labeled question $\boldsymbol{h}^{q, 1}$ and unlabeled question $\boldsymbol{h}^{u_n}$ via respective fully connected layers.
Then the representations are activated by $\tanh$:
\begin{equation}
\begin{split}
h_t^{q, r}=\tanh (W^r h_t^{q, 1}+b^r )  \\
h_v^{u_n, k}=\tanh (W^k h_v^{u_n}+b^k ) ,
\end{split}
\end{equation}
where $W^r$, $b^r$, $W^k$ and $b^k$ are trainable weights. 
The $t$-th word in the labeled question first obtain the attention weight for the $v$-th word in the $n$-th unlabeled question via a dot product.
Then the weights are normalized by a softmax function:
\begin{equation}
\begin{split}
\alpha_{t, v}^{q, u_n}=\frac{\exp{\big((h_t^{q, r})^T h_v^{u_n, k}\big)} }{\sum_{v'=1}^{T_{u_n}} \exp{\big((h_t^{q,r})^T h_{v'}^{u_n, k}\big)} } .
\end{split}
\end{equation}
This is the level 1 attention weights.
Let $h_t^{q, u_n}$ denote the side information of the $t$-th word in the labeled question for the $n$-th unlabeled question (representation after the first-level attention).
It is the weighted sum over all words in the $n$-th unlabeled question.
\begin{equation}
\begin{split}
h_t^{q, u_n}=\sum_{v=1}^{T_{u_n} } \alpha_{t, v}^{q, u_n} h_v^{u_n, k}.
\end{split}
\end{equation}
Later, we have a level 2 attention over different unlabeled questions. 
Again we first transform the side information of the $t$-th word for each unlabeled question: 
\begin{equation}
\begin{split}
h_t^{q, u_n, k'}=\tanh (W^{k'} h_t^{q, u_n} + b^{k'} ) .
\end{split}
\end{equation}
Then the level 2 attention weights are again obtained via dot products normalized by a softmax function:
\begin{equation}
\begin{split}
\alpha_t^{q, u_n}=\frac{\exp{\big((h_t^{q, r})^T h_t^{q, u_n, k'}\big)} }{\sum_{n'=1}^{|U|} \exp{\big((h_t^{q,r})^T h_t^{q, u_{n'}, k'} \big)} } .
\end{split}
\end{equation}
And finally the side information vector for the $t$-th word in the labeled question (representation after level 2 attention) is:
\begin{equation}
\begin{split}
s_t^q=\sum_{n=1}^{|U|} \alpha_{t}^{q, u_n} h_t^{q, u_n, k'}.
\end{split}
\end{equation}

Lastly, we concatenate $s_t^q$ with $h_t^{q,1}$ as the representation of the $t$-th word in the question: $h_t^{q,2}=h_t^{q,1} \oplus s_t^q$. 

\subsection{Sequence Labeling}
After obtaining the representation of the labeled question with side information, we feed $\bm{h}^{q,2}$ into another BLSTM layer. 
So we have two LSTM layers for the labeled question, which is similar to the stacked BLSTM \cite{el1995hierarchical} (S-BLSTM). 
We use S-BLSTM to obtain better sequence representation. 
Then we have $\boldsymbol{h}^{q,3}$ for the labeled question sequence. 
We reduce the dimension of $h_t^{q,3}$ to the size of the label set via a fully connected layer:

\begin{equation}
\begin{split}
c_t^q =W h_{t}^{q,3} +b ,
\end{split}
\end{equation}
where $c_t^q \in \mathbb{R}^{|L|}$. We output the probability distribution over labels $L$ for the $t$-th question word via a softmax function: 
\begin{equation}
\begin{split}
p^q(\hat{y}_t=l|\mathbf{x}^q, \mathbf{x}^{u_{1:|U|}}; \Theta) & =\frac{\exp(c_{t, l}^q)}{\sum_{l' \in L}\exp(c_{t, l'}^q)} ,
\end{split}
\end{equation}
where $\Theta$ represents all trainable parameters, including parameters in LSTM cells and word embeddings. 
Finally, we optimize the cross entropy loss function over the training dataset:
\begin{equation} \label{eq:opt}
\begin{split}
J(\Theta)=-\sum_m^{|M|} \sum_t^{|T_q|} \sum_{l \in L} y_{t, l}^{(m)} \log p^q(\hat{y}_t^{(m)}=l| \mathbf{x}^q, \mathbf{x}^{u_{1:|U|}} ; \Theta) ,
\end{split}
\end{equation}
where $M$ represents all the training examples. 
$y_{t, l}^{(m)} \in \{0, 1\}$ is the ground truth for the $t$-th question word and label $l$ in the $m$-th training example. 
We leverage Adam optimizer \cite{kingma2014adam} to optimize the whole network.
We set the learning rate as 0.001 and keep other parameters the same as the original paper. 
We set the dropout rate to 0.2. The batch size is set to 256.

%During testing, the prediction for each position in the labeled question given unlabeled questions is computed as:
%\begin{equation} \label{eq:pred}
%\begin{split}
%\hat{y_t}=\operatorname*{arg\,max}_{l \in L} p^q(\hat{y}_t=l| \mathbf{x}^q, \mathbf{x}^{u_{1:|U|}}; \Theta) .
%\end{split}
%\end{equation}

\section{Experimental Result}
\label{sec:exp}

\begin{table}
\footnotesize
\centering
\scalebox{1}{
\begin{tabular}{ l | c | c }
\hline
Product & QA & \% of QAs with Functions \\
\hline
DSLR & 327 & 20.18 \\
E-Reader & 271 & 31.37 \\
Speaker & 153 & 30.72 \\
Tablet & 329 & 42.86 \\
Cellphone 1 & 170 & 57.65 \\
Cellphone 2 & 330 & 41.82 \\
Laptop 1 & 297 & 18.86 \\
Laptop 2 & 425 & 54.59 \\
Netbook & 199 & 44.72 \\
TV & 306 & 46.41 \\
TV Console & 183 & 54.1 \\
Gaming Console & 212 & 70.28 \\
Apple Watch & 331 & 28.1 \\
\hline
VR Headset & 444 & 76.13 \\
Stylus & 266 & 71.05 \\
Micro SD Card & 283 & 81.27 \\
Mouse & 259 & 66.02 \\
Tablet Stand & 214 & 88.79 \\
\hline
Total & 4999 & 51.07\\
\hline 
\end{tabular}
}
\caption{Statistics of 18 labeled products. QAs: number of QA pairs; \% of QAs with Functions: percentage of QA pairs containing function needs.}
\label{table:dataset}
\end{table}

\subsection{Corpus Annotation, Analysis, and Preprocessing}
\label{sec:data}
We crawled about 1 million QA pairs from the pages of products in the electronics department from Amazon as the training corpus for skip-gram model \cite{mikolov2013distributed} to obtain word embedding matrix $W_e$.

We further annotated a subset of 4999 QA pairs from 18 products for model training and testing.
The basic statistics of the corpus is
shown in Table \ref{table:dataset}.
The corpus is labeled by 3 annotators independently. The general annotation guidelines are as follows:
\begin{enumerate}
\item only yes/no QAs should be labeled;
\item a function expression is labeled as a function target with an optional function verb; 
\item a function target can be specific entities (e.g., ``iPhone''), general entities like ``video'' or service providers like ``AT\&T'';
\item a function target should be labeled as token spans containing nouns, adjectives, or model numbers (e.g., ``Samsung micro SD EVO''); 
\item expressions about specific aspects or accessories are not considered as function expressions. This is because aspects or accessories are not closely related to the functionality of the product as a whole;
\item nouns that are subjective are not regarded as function target (e.g., the word ``need'' in ``Can it fit my need ?'');
\item the optional function word can be a verb (e.g., ``produce'' in ``produce music'') or its noun form (e.g., ``production'' in ``music production''); we also include the adjunct word (e.g., ``with'' in ``work with iPhone'') for extrinsic function expression;
\item some function expression does not have function word, e.g., ``Does Skype ok on this?'';
\end{enumerate}

All annotators initially agreed on their annotations
(same function targets and function words)
on 81\% of all QA pairs.
Disagreements are then resolved to reach final consensus annotations.

We observe that accessories (the last 5 products) have a higher percentage of the function need related questions than those of main products (the first 13 products).
This is expected since one accessory may work with multiple devices and thus have more functions. 

The annotated corpus is preprocessed using Stanford CoreNLP \footnote{http://stanfordnlp.github.io/CoreNLP/}.
We have the following steps: sentence segmentation, tokenization, POS-tagging, lemmatizing and dependency parsing. 
The last 3 steps provide features for the Conditional Random Fields (CRF) \cite{lafferty2001conditional} baseline. 

We also select the most similar 5 unlabeled questions under the same category as the labeled question returned by ElasticSearch\footnote{www.elastic.co}, as the question bank. 

We only perform sentence segmentation and tokenization on these unlabeled questions to save preprocessing time. 
Lastly, multiple sentences in both labeled and unlabeled questions are concatenated together. 
We set the maximum length of a question to be 40. 
This covers 99.5\% labeled questions in full length. 

After preprocessing, one example contains a labeled question, 5 unlabeled questions, and one labeled answer.
We shuffle all examples and select 70\% for training, 10\% for validation and 20\% for testing. 
The validation set is used to avoid overfitting on the training data.

\subsection{Baselines}

\begin{table}
%\footnotesize
\centering
\scalebox{0.95}{
\begin{tabular}{ l | c | c | c }
\hline
Method & $\mathcal{P}$ & $\mathcal{R}$ & $\mathcal{F}_1$ \\
\hline
CRF & 0.798 & 0.611 & 0.692 \\
S-BLSTM & \textbf{0.844} & 0.673 & 0.749 \\
SAN (-) BLSTM2 & 0.83 & 0.7 & 0.759 \\
SAN & 0.839 & \textbf{0.721} & \textbf{0.776} \\
\hline 
\end{tabular}
}
\caption{Different methods for Function Need Recognition (FNR) in precision, recall and F1-score.}
\label{table:comparison}
\end{table}

We compare the following baselines with SAN:
\begin{enumerate}
\item \textbf{CRF}: 
We use Mallet\footnote{http://mallet.cs.umass.edu/} as the CRF implementation.
We train a CRF model using exactly the same training data as the proposed method. 
We use the following manually created features: 
\begin{enumerate} 
\item the words within a 5-word window;
\item the POS tags within a 5-word window;
\item the number of characters;
\item binary indicators (camel case, digits, dashes, slashes and periods);
\item dependency relations for the current word obtained via dependency parsing. 
\end{enumerate}
We use CRF as a baseline to show the performance of a non-deep learning method.

\item \textbf{S-BLSTM}: 
This baseline is a traditional S-BLSTM with 2 layers (by removing the bank attention from SAN). 
It is a supervised baseline. 
We use this baseline to show that using purely supervised data is not good enough. 
Unlabeled data can help to improve the performance.

\item \textbf{SAN (-) BLSTM2}: 
This baseline does not have the second layer of BLSTM for the labeled question. 
We use this baseline to show that S-BLSTM works better for our problem. We use 5 unlabeled questions in both this baseline and SAN.
\end{enumerate}

\textbf{Result Analysis} 
From Table \ref{table:comparison}, we can see that the proposed SAN framework performs the best on F1-score. 
Although CRF is a non-deep learning model, its precision is not bad since we use dependency relations as features.
However, the recall of CRF is very low since it can only train weights on words appear in the training data. 
All deep learning models have better recalls than CRF.
S-BLSTM has the best precision as it is trained using only the training data. 
However, its recall is relatively low.
It still suffers the problem that training data can not further tune embeddings of words not appeared in the training data. 
SAN (-) BLSTM2 shows that the additional BLSTM layer is effective in learning better representations.
Lastly, SAN significantly improves the recall by further adjusting the weights for different unlabeled questions. 
It only loses 0.5\% on precision compared that with S-BLSTM.

\section{Related Work}
\label{sec:rw}
Both data mining and natural language processing communities study sentiment analysis on products\cite{hu2004mining,mcauley2013hidden,McAPanLes15,liu2015sentiment,shu-xu-liu:2017:Short}. 
However, Product Community Question and Answering (PCQA) only draws attention in recent years \cite{McAYan16,liuretrieving}.
PCQA is studied as a relevance ranking problem in \cite{McAYan16,liuretrieving}.
Given a question, they retrieve relevant reviews to augment existing answers. 
Instead, we observe that PCQA also contains valuable fine-grained information for extraction. 
Product function needs are an important type of such information. 
Functions may contain both intrinsic functions and extrinsic functions \cite{Xu2018pro}.
Extrinsic functions are closely related to complementary products (taking whether one product can work with another as a function) \cite{xu2016CER,xu2016pcqa,xu2017supervised}.
But we observe that from the perspective of functionality, how two products can work together is also important.
For example, ``install Windows 10'' and ``run Windows 10'' are two different functions.

Although CNN \cite{kim2014convolutional,shu-xu-liu:2017:EMNLP2017} and Long Short-Term Memory (LSTM) \cite{hochreiter1997long} are both used in NLP tasks, LSTM is more commonly used in sequence labeling \cite{greff2015lstm,lample2016neural}.
Attention mechanism is popular in image recognition \cite{NIPS2010_4089,denil2012learning}.
It is later used in natural language processing \cite{hermann2015teaching,xu2015show}. 
However, attention mechanism is only used in supervised settings. 
We adapt attention for a semi-supervised setting \cite{zhu2005semi}.
Traditional semi-supervised learning uses unlabeled data as training examples \cite{rasmus2015semi} directly.
Instead, we use unlabeled data as side information for labeled examples. 

\section{Conclusion}
In this paper, we propose the task of Function Need Recognition (FNR), which is to identify function needs queried by customers.
We leverage a Semi-supervised Attention Network (SAN) to solve this problem by leveraging unlabeled data as attended side information.
Experiments demonstrate that the SAN is better than a number of baselines.

\section*{Acknowledgment}
This work is supported in part by NSF through grants IIS-1526499, and CNS-1626432, and NSFC 61672313.

\bibliographystyle{IEEEtran}
\bibliography{IEEE}
% that's all folks
\end{document}